# Training-Free Style Consistent Image Synthesis with Condition and Mask Guidance in E-Commerce

Guandong Li[1*]

1.*Suning, Xuanwu, Nanjing, 210042, Jiangsu, China.

*Corresponding author(s). E-mail(s): leeguandon@gmail.com

*Abstract*: Generating style-consistent images is a common task in the e-commerce field, and current methods are largely based on diffusion models, which have achieved excellent results. This paper introduces the concept of the QKV (query/key/value) level, referring to modifications in the attention maps (self-attention and cross-attention) when integrating UNet with image conditions. Without disrupting the product's main composition in e-commerce images, we aim to use a train-free method guided by pre-set conditions. This involves using shared KV to enhance similarity in cross-attention and generating mask guidance from the attention map to cleverly direct the generation of style-consistent images. Our method has shown promising results in practical applications.

1. Introduction

   Diffusion-based text-to-image models have had a lasting impact on the e-commerce field. Generating style-consistent images is a common task in e-commerce. Merchants create images with similar styles based on existing ones. These images are then used for platform placements.

   Generating style-consistent images involves incorporating image conditions. Common text-to-image models, such as Stable Diffusion, use three main methods for integrating external condition information. The first method involves concatenating with latents sampled during image generation, as applied in the original Stable Diffusion webui img2img method. The second method interacts through cross attention in the Spatial Transformer of the UNet module, similar to text features, with concatenation methods including before cross attention and decoupled cross attention. The third method is similar to ControlNet. We chose the second method, specifically the ip-adapter, as our foundational approach.

   One of the main methods for controlled generation in Stable Diffusion involves performing attention map operations at the QKV level. This can be broadly categorized into two dimensions: alignment between text prompts and generated images, and alignment between image prompts and generated images. In Stable Diffusion, the QKV in the UNet's self-attention and cross-attention modules is editable. Editing QKV directly affects the generated image. After Q and K, softmax produces the attention map, and the product of the attention map and V serves as the feature passed on. K and V as text embeddings establish a mapping with Q, i.e., the generated image. By manipulating KV, we can control the generation of objects corresponding to the text. Similarly, image alignment can involve image features obtained from feature extractors or image latents acquired through inversion.

   In e-commerce, generating stylistically similar images faces two main issues: 1. The similarity generation in e-commerce aims to preserve the product's main composition while maintaining a similar style and ensuring diversity without being overly uniform; 2. It has been observed that text prompts often have a strong influence, and unreasonable text prompts can significantly disrupt the style. To address these issues, we designed similarity image generation using condition and mask guidance, while aiming for a train-free approach. Condition guidance involves inputting two sets of text prompts:

the first set is user-defined, and the second set consists of pre-configured text conditions based on the first set. We perform batch-2 forward inference, where in the decoupled cross-attention, we use the key and value from batch 1 to replace those in batch 2. This strategy helps ensure consistent style under diverse text prompt guidance. Mask guidance involves visualizing the attention map formed by softmax. We find that there is a clear mapping between both text prompts and image prompts with the final generated image. To maintain the structure of the final image relatively unchanged, we enhance this guidance by thresholding the attention map to create a threshold map, and then multiplying the threshold map with V to better guide image generation.

Our contributions include: 1. Introducing a train-free approach for e-commerce similarity image generation; 2. Incorporating condition and mask guidance.

2. Related works

IP-adapter. IP-adapter is a work with significant impact. It introduces an image-conditioned branch, extracting image features through an image encoder and aligning modalities via a linear layer. Text and image features use decoupled cross-attention, with an additional cross-attention layer specifically for image features added to the diffusion model's text embedding. During training, only the parameters of the new cross-attention layer are updated, while the original UNet model remains frozen, resulting in an IP-adapter with only 22M parameters. It can generate images with similar styles based on an input image and serves as the foundational framework for this paper.

Controlnet reference. Controlnet reference is a highly creative train-free method requiring two rounds of diffusion. In the first round, the VAE encodes the reference image into latent space, then interpolates between the latent features of the reference and randomly initialized features. In SD1.5, there are 16 transformer blocks in the UNet, generating and caching 16 hidden features during the first diffusion's forward pass. In the second forward diffusion, the self-attention mechanism uses the feature cache from the corresponding transformer block layers of the first diffusion as a reference. There are various self-attention variants based on performance. This inspired the approach of using batch-2 inference for cross-attention QKV operations in this paper.

QKV level operations. This aspect of work includes Masactrl[5], which transforms the existing self-attention in diffusion models into mutual self-attention, enabling it to query relevant local content and textures from the source image to maintain consistency. Cross-image attention[6] involves a pair of images: one depicting the target structure and the other specifying the desired appearance. The cross-image attention combines queries from the structural image with the keys and values from the appearance image. When applied during the denoising process, this operation generates images that blend the desired structure and appearance by utilizing established semantic correspondences. StyleAligned[7] demonstrates that minimal attention-sharing operations from each generated image to the first image in the batch during the diffusion process can produce a set of images with a consistent style. Prompt-to-Prompt[8] focuses on image editing based on editing instructions while ensuring that the edits do not cause significant changes to the image. This is achieved by injecting cross-attention maps into the diffusion model's diffusion process to control which tokens of the prompt text the pixels focus on during diffusion, thereby editing the image. Self-guidance[9] ensures consistency of the same object across different generated images by adding loss control between the attention maps corresponding to text in two inference rounds.

3. Method

In this section, we will provide a detailed description of the proposed train-free style-consistent image design framework, as shown in Figure 1. Given an original style image $I_s$, our goal is to synthesize an image $I$ that conforms to the style of $I_s$ under a newly provided prompt $P$. The newly synthesized target image $I$ should have the same spatial distribution as $I_s$ and should retain the style and subject composition of $I_s$.

The core idea is to extend the specified prompt $P$ and image into a batch mode, using KV reuse to query semantically similar content from the original image. By performing mask extraction on the attention map formed after softmax, we guide image generation with the mask to ensure that the subject composition remains unchanged.

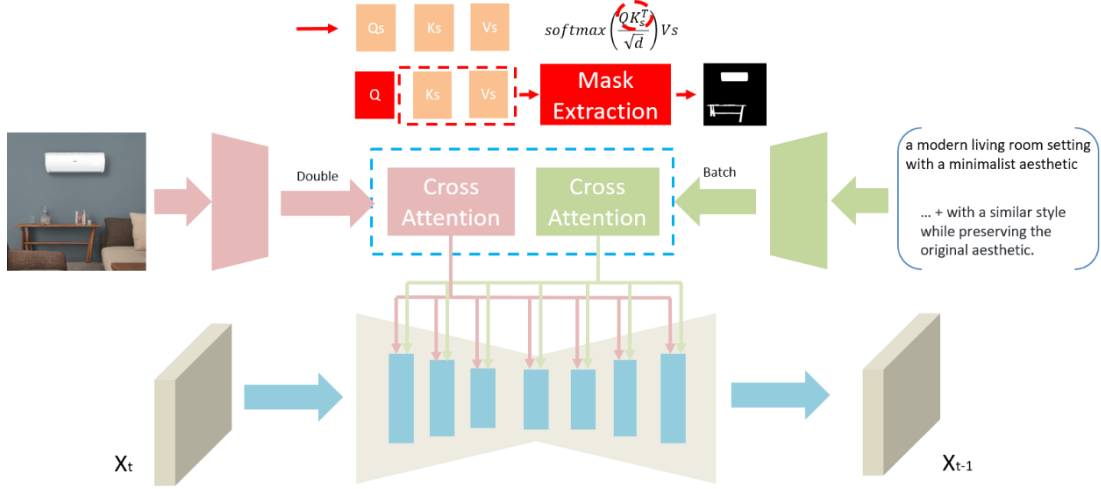

Figure 1: Consistent Image Generation Architecture under Condition and Mask Guidance

3.1 Base Architecture

Our base architecture uses the IP-Adapter framework, and the method itself is train-free, which means condition and mask guidance can be seamlessly integrated into any IP-Adapter series algorithms. In the diagram above, we continue with the decoupled cross-attention of IP-Adapter. The projection weights for the key and value of the pre-trained cross-attention are trained to adapt to text features, so merging image and text features and then applying cross-attention only aligns image and text features but may miss some image-specific information, potentially resulting in coarse-grained controllable generation based solely on the reference image. Therefore, decoupled cross-attention is also very important in our design.

3.2 QKV level

The core design work of this paper focuses on the condition and mask guidance modules. Essentially, both of these modifications pertain to the QKV level. Changes at the QKV level are highly promising in the optimization of stable diffusion, particularly in the self-attention and cross-attention mechanisms of the UNet module. Here, Q represents the learnable object, which is the image feature, while KV typically represents text features. The purpose of cross-attention in Stable Diffusion's UNet is to encode text information as a condition into the generation process. According to the attention formula [10], the essence of the attention operation is to aggregate V based on the relationship between Q and K. Therefore, text information needs to be used as KV in this context. If swapped, it would be challenging to integrate text information into the image generation process, making it difficult to produce results that match the text. This paper introduces the concept of the QKV level.

### 3.2.1 Condition guidance

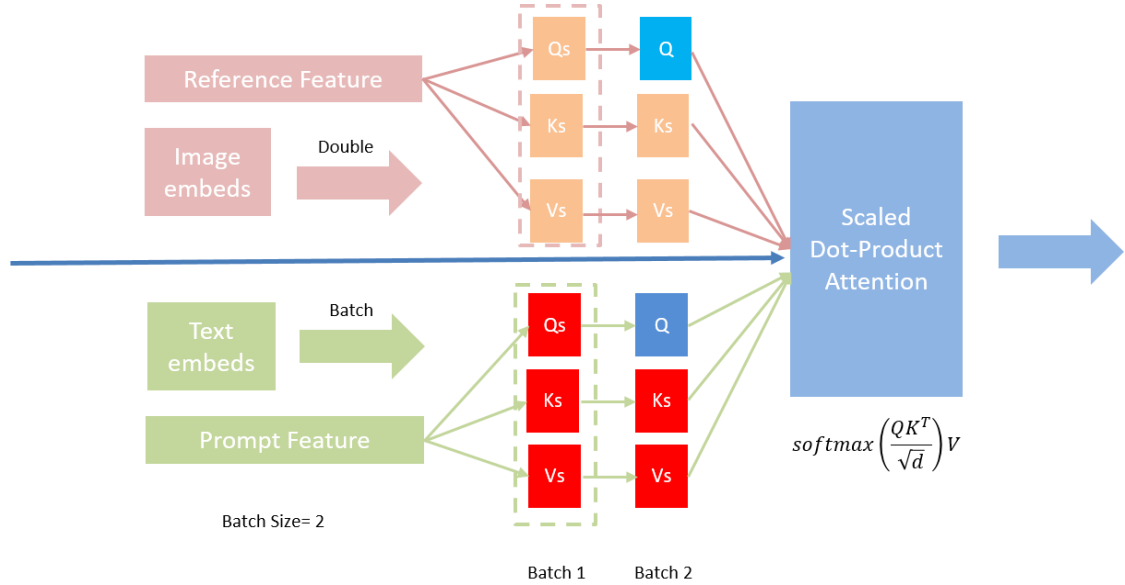

Figure 2: Structure diagram of condition guidance

As shown in Figure 2, we use different processing methods for image and text inputs. For the image side, after extracting features, we double the dimensions to form a batch size of 2. On the text side, we embed two sets of text prompts, namely our preset text condition and the prompt $P$, resulting in a feature dimension with batch size 2. In the decoupled cross-attention, we input both sets of features simultaneously. For the image and text, we replace the batch 2 K and V with the batch 1 K and V, while Q remains from batch 2. The content features come from $I_s$ and $P$, thus forming a shared KV. Our inspiration comes from Masactrl and Controlnet reference. In Controlnet reference, there are two diffusion processes, with the second diffusion process using the KV from the first diffusion process in self-attention. In Masactrl, self-attention is transformed into multi self-attention. The sharing of KV indicates that the target image focuses only on itself as well as the reference image and text, producing a consistent style while maintaining diversity.

### 3.2.2 Mask guidance

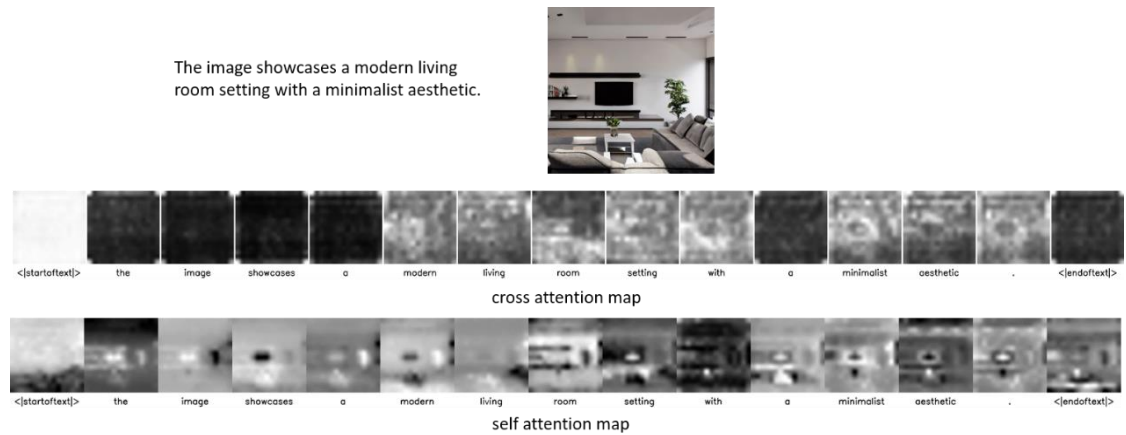

Figure 3: Mapping relationship between cross attention, self-attention map, and text prompt

While maintaining style consistency, we also aim to keep the consistency of the e-commerce subject composition. As shown in Figure 3, both the cross attention map and the self-attention map have a clear mapping relationship with the text prompt. The self-attention map already contains most of the shape and structure information, while cross attention, being weaker when text is embedded, guides less effectively. Therefore, we use the cross attention map to create a mask, which is then used to guide image generation. In this study, we only applied mask guidance to the attention map on the text prompt side. It is possible to guide the attention map on the image prompt side as well, but it requires careful handling. The activation of the text prompt's attention map and its mapping to the image generation area are clear, and the overall activation in mask guidance is relatively small, thus having minimal impact on the global image's semantic integrity. However, the situation is different for the image prompt side. Since the algorithm itself is train-free, directly using a mask image for guidance can introduce significant deviations. We are still searching for better mask guidance methods.

4. Results

As shown in Figure 4, it demonstrates the results of our condition and mask guidance under reference images. In most cases, it preserves the main composition structure of the original e-commerce reference images, and it performs well even with minimal or no prompt input.

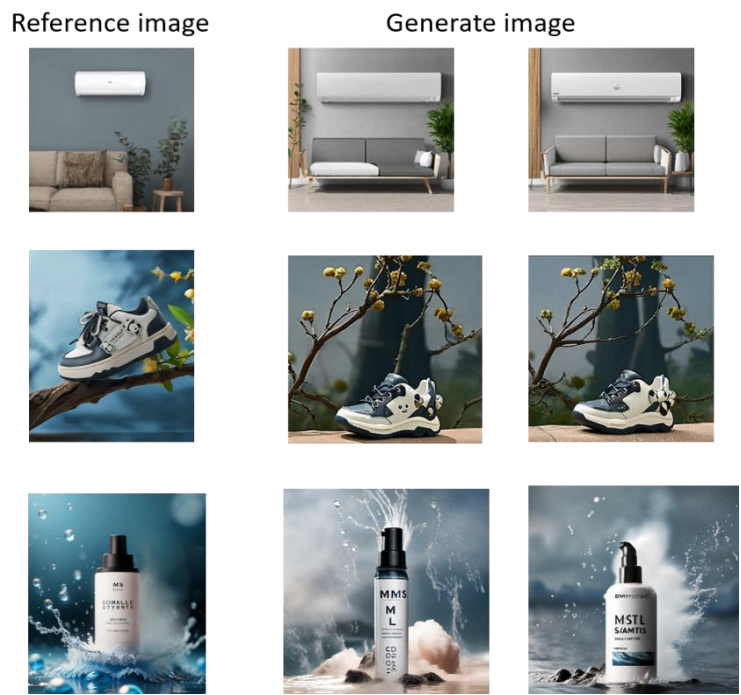

Figure 4: Model results

5. conclusion

In this paper, for style-consistent image generation in e-commerce scenarios, we introduce condition guidance and mask guidance based on the IP-adapter structure. This approach achieves good results without compromising the main composition or diversity of the images. By using preset conditions, KV sharing, and enhancing similarity in cross attention through KV replacement, along with generating mask guidance from the attention map, we effectively guide the generation of style-consistent images. Our train-free algorithm shows promising application prospects and can be integrated into the IP-adapter series. We hope this paper will serve as a technical report opening new approaches for style-consistent image generation at the QKV layer in the e-commerce field.